\documentclass{article}

\usepackage{arxiv}

\usepackage[utf8]{inputenc} 
\usepackage[T1]{fontenc}    
\usepackage{hyperref}       
\usepackage{url}            
\usepackage{booktabs}       
\usepackage{amsfonts}       
\usepackage{nicefrac}       
\usepackage{microtype}      
\usepackage{lipsum}		
\usepackage{graphicx}
\usepackage{natbib}
\usepackage{doi}

\usepackage{algorithm}
\usepackage{algpseudocode}
\usepackage{amsmath}

\title{Efficient Second-Order Neural Network Optimization via Adaptive Trust Region Methods}

\author{ \href{https://orcid.org/0000-0002-4363-2177}{\includegraphics[scale=0.06]{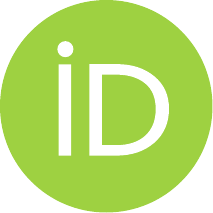}\hspace{1mm}James Vo}\thanks{Anh-Dung Vo} \\
	DocumentAI Team\\
	AGILESODA INC.\\
	Seoul, 06149, South Korea \\
	\texttt{anhdungitvn@agilesoda.ai} \\
}

\renewcommand{\shorttitle}{Efficient Second-Order Neural Network Optimization via Adaptive Trust Region Methods}

\hypersetup{
pdftitle={Efficient Second-Order Neural Network Optimization via Adaptive Trust Region Methods},
pdfsubject={q-bio.NC, q-bio.QM},
pdfauthor={David S.~Hippocampus, Elias D.~Striatum},
pdfkeywords={First keyword, Second keyword, More},
}

\begin{document}
\maketitle

\begin{abstract}

Second-order optimization methods offer notable advantages in training deep neural networks by utilizing curvature information to achieve faster convergence. However, traditional second-order techniques are computationally prohibitive, primarily due to the large matrix inversions and high memory demands they require. While adaptive trust-region methods have been developed to mitigate these issues, their performance is often hindered by conservative estimates of key parameters, such as the Lipschitz constant of the Hessian, resulting in suboptimal outcomes. In this paper, we introduce SecondOrderAdaptiveAdam (SOAA), a novel optimization algorithm designed to overcome these limitations. SOAA approximates the Fisher information matrix using a diagonal representation, reducing computational complexity from \(O(n^{2})\) to \(O(n)\), thereby making it suitable for large-scale deep learning models, including large language models (LLMs). Additionally, the algorithm integrates an adaptive trust-region mechanism that dynamically adjusts the trust region size based on observed loss reduction, ensuring both robust convergence and computational efficiency. We empirically demonstrate that SOAA achieves faster and more stable convergence compared to first-order optimizers, such as Adam, under similar computational constraints. However, the diagonal approximation of the Fisher information matrix may be less effective in capturing higher-order interactions between gradients, suggesting potential areas for further refinement and future research.

\end{abstract}


\section{Introduction}

Second-order optimization methods offer significant advantages over their first-order counterparts by incorporating curvature information from the Hessian matrix or the Fisher information matrix. This enhanced gradient estimation enables faster convergence, particularly in complex optimization landscapes commonly encountered in deep learning. However, traditional second-order methods are often computationally prohibitive due to the large matrix inversions required, especially for high-dimensional neural network models (\cite{hamad2024consistentlyadaptivetrustregionmethod, curtis2022worstcasecomplexitysqpmethod, na2022hessianaveragingstochasticnewton}).

In this paper, we propose SOAA, a second-order optimization algorithm that combines the advantages of curvature-aware methods with the efficiency of first-order optimizers like Adam. Traditional second-order methods often suffer from high computational complexity due to the need for full matrix computations and inversions, making them impractical for large-scale deep learning models. SOAA addresses this challenge by employing a diagonal approximation of the Fisher information matrix, reducing the complexity to \(O(n)\), and enabling second-order optimization with a memory footprint similar to first-order methods.

Additionally, SOAA incorporates an adaptive trust-region mechanism that dynamically adjusts the permissible step size based on the observed versus expected loss reduction. This mechanism prevents large, unstable updates while allowing the optimizer to adapt more aggressively when beneficial, ensuring fast and stable convergence.

Our methodology focuses on three key innovations:

1. \textbf{Diagonal Fisher Information Matrix Approximation}:
   SOAA approximates the Fisher information matrix using a diagonal representation. This approach captures second-order curvature information while avoiding the computational burden of a full matrix inversion, reducing complexity from \(O(n^{2})\) to \(O(n)\).

2. \textbf{Adaptive Trust Region}:
   Inspired by trust-region methods, SOAA adjusts the trust region dynamically during training. If the observed loss reduction exceeds expectations, the trust region expands, enabling larger parameter updates. Conversely, if the loss reduction is smaller than expected, the region contracts, ensuring more conservative updates to maintain stability.

3. \textbf{Efficient Gradient and Moment Updates}:
   SOAA retains the adaptive moment estimation of Adam, updating both the first and second moments of the gradients. This ensures that momentum is maintained, while also benefiting from second-order curvature information for more precise updates.

We validated the effectiveness of SOAA through experiments across various models and datasets. SOAA demonstrated faster convergence and greater stability compared to standard optimizers such as Adam, AdamW, and memory-efficient variants like \href{https://huggingface.co/docs/bitsandbytes/main/en/optimizers}{8bit-Adam} and GaLore (\cite{zhao2024galorememoryefficientllmtraining}). Although SOAA requires more memory, its performance benefits make it suitable for large-scale deep learning tasks where optimization speed and robustness are critical.

Our method builds upon previous work in trust-region methods, such as those by Hamad et al. (\cite{hamad2024consistentlyadaptivetrustregionmethod}), who introduced consistently adaptive trust-region methods that dynamically adjust based on local curvature without requiring conservative estimates of problem properties like the Lipschitz constant. SOAA extends these ideas to large-scale neural network training, maintaining efficiency while providing the accuracy and stability of second-order methods.

In conclusion, SOAA offers a powerful solution for scalable second-order optimization in deep learning, effectively balancing computational efficiency, memory usage, and convergence performance. Future work will explore further enhancements to the Fisher information approximation, such as incorporating off-diagonal elements or low-rank methods, to further improve scalability and performance in even larger models.

\section{Methodology}
\label{sec:methodology}

\subsection{Second-Order Approximation Using Fisher Information}

The Fisher information matrix provides a second-order approximation of the loss function, encapsulating curvature information critical for effective parameter updates. Directly computing the full Fisher matrix is computationally expensive, especially for large-scale deep learning models. Inspired by work on trust-region methods and second-order approaches (\cite{hamad2024consistentlyadaptivetrustregionmethod}), we decompose the Fisher matrix into the outer product of gradients and a diagonal covariance matrix. This diagonal approximation allows us to maintain second-order information while reducing the computational complexity to \(O(n)\).

This approach also mitigates the memory overhead typically associated with second-order methods. By approximating the Fisher matrix as a diagonal matrix, SOAA avoids the need for full matrix inversion, which is particularly costly in high-dimensional spaces. This approximation is particularly effective in deep learning applications, where the dimensionality of the parameter space can easily reach billion.

\subsubsection{Adaptive Trust Region}
To address the limitations of fixed learning rates, SOAA incorporates an adaptive trust region. The trust region determines the extent to which parameters can be adjusted without incurring significant increases in the loss function. Building on trust-region methods (\cite{hamad2024consistentlyadaptivetrustregionmethod}), we adaptively adjust the trust region size based on the observed versus anticipated loss reduction. If the observed loss reduction is greater than expected, we expand the trust region to allow for more aggressive updates. Conversely, if the observed reduction is less than anticipated, we shrink the trust region to prevent overshooting during optimization.

This adaptive trust-region mechanism ensures that SOAA remains stable during training, preventing large oscillations in the parameter updates while still allowing for dynamic adjustments in response to changes in the loss landscape.

\subsubsection{Computational Efficiency}
By using a diagonal approximation of the Fisher information matrix, SOAA achieves a computational complexity of \(O(n)\), similar to that of first-order methods such as Adam. This reduction in complexity is crucial for the scalability of second-order methods in deep learning, where models often have millions or even billions of parameters. Element-wise vector operations are used to compute parameter updates, avoiding the need for expensive matrix operations like inversion, which can become a bottleneck in large-scale optimization tasks.

The adaptive trust-region mechanism also ensures that computational resources are used efficiently. By dynamically adjusting the trust region size, SOAA balances the need for precise parameter updates with the computational cost of each iteration, ensuring that updates are made within a computationally feasible time frame.

\subsubsection{How It Works}

\begin{algorithm}
\caption{SOAA Optimizer}
\label{algorithm:SecondOrderAdaptiveAdamOptimizer}
\begin{algorithmic}[1]
\State \textbf{Input:} Parameters $\theta$, gradients $g$, learning rate $\alpha$, betas $(\beta_1, \beta_2)$, gamma $\gamma$, epsilon $\epsilon$, weight decay $\lambda$, total steps $T$
\State Initialize $m = 0$, $v = 0$, $s = 0$, trust region scaling $dt = 1$, loss average $l_{\text{avg}} = 0$, predicted reduction $pr = 0$
\For{each iteration $t = 1, 2, ..., T$}
    \State Compute gradient $g_t$ for current parameters $\theta_t$
    \State Update first moment estimate: $m_t = \beta_1 m_{t-1} + (1 - \beta_1) g_t$
    \State Update second moment estimate: $s_t = \beta_2 s_{t-1} + (1 - \beta_2) g_t^2$
    \State Bias correction:
    \begin{align*}
        \hat{m}_t &= \frac{m_t}{1 - \beta_1^t} \\
        \hat{s}_t &= \frac{s_t}{1 - \beta_2^t}
    \end{align*}
    \State Compute Fisher information approximation: 
    \begin{align*}
        F_t = \left( 1 + \frac{\sum \hat{m}_t^2}{\sum (\hat{v}_t + \epsilon)} \right) \hat{v}_t
    \end{align*}
    \State Compute trust region scaling:
    \begin{align*}
        \text{trust\_region\_scale} = \max(dt \cdot F_t, \sqrt{\hat{s}_t})
    \end{align*}
    \State Update adjusted gradient:
    \begin{align*}
        g_{\text{adjusted}} = \frac{\hat{m}_t \cdot dt}{\text{trust\_region\_scale} + \epsilon}
    \end{align*}
    \If{weight decay $\lambda$ is not zero}
        \State Apply weight decay: $\theta_t = \theta_t - \alpha \lambda \theta_t$
    \EndIf
    \State Update parameters: $\theta_t = \theta_t - \alpha g_{\text{adjusted}}$
    \If{loss is available}
        \State Update loss average: $l_{\text{avg}} = \beta_1 l_{\text{avg}} + (1 - \beta_1) \cdot l_t$
        \State Bias corrected loss average: $\hat{l} = l_{\text{avg}} / (1 - \beta_1^t)$
        \State Update trust region scaling $dt$: 
        \begin{align*}
            dt = \min\left(\max\left(\frac{(\hat{l} - l_t)}{\max(pr, \epsilon)} \cdot dt, (1 - \gamma)^{\frac{t-1}{T}}\right), 1 + \gamma^{\frac{t-1}{T}}\right)
        \end{align*}
        \State Update predicted reduction:
        \begin{align*}
            pr = \left((\hat{m}_t \cdot g_{\text{adjusted}}) - 0.5 \cdot (\hat{v}_t \cdot g_{\text{adjusted}}^2)\right) \cdot \alpha
        \end{align*}
    \EndIf
\EndFor
\end{algorithmic}
\end{algorithm}

The core mechanism of the SOAA optimizer combines the efficiency of first-order methods, like Adam, with second-order information by employing a scalable approximation of the Fisher information matrix. The SOAA optimizer is formally presented in Algorithm \ref{algorithm:SecondOrderAdaptiveAdamOptimizer}, which outlines its key steps. This approach ensures stability and fast convergence while maintaining computational efficiency.

The optimizer works by following several key steps:

1. \textbf{Gradient and Moment Calculation}:
   At each optimization step, the gradients of the parameters are computed based on the current loss. Similar to the Adam optimizer, we maintain running estimates of both the first moment (mean of the gradients) and the second moment (uncentered variance of the gradients). These moment estimates are updated in each iteration using exponential decay rates, controlled by the hyperparameters \(\beta_1\) and \(\beta_2\), respectively. This mechanism allows for adaptive momentum in gradient updates.

2. \textbf{Fisher Information Approximation}:
   The optimizer approximates the Fisher information matrix by leveraging a diagonal covariance matrix representation. Rather than computing the full Fisher matrix, which is computationally expensive, SOAA uses the outer product of the gradient vector combined with a diagonal approximation to capture second-order curvature information. This approximation reduces the complexity to \(O(n)\), making the optimizer scalable for large neural networks.

3. \textbf{Adaptive Trust Region}:
   A key innovation in SOAA is the adaptive trust region mechanism. The trust region controls how much the parameters can be updated in a single step without causing instability. Based on the observed loss reduction, the trust region is dynamically adjusted. If the observed reduction is larger than expected, the trust region expands, allowing more aggressive updates. Conversely, if the observed reduction is smaller than expected, the trust region contracts, ensuring more cautious updates. This adaptive behavior prevents the optimizer from overshooting or getting stuck in local minima.

4. \textbf{Parameter Updates}:
   After computing the moments and adjusting the trust region, the parameters are updated. The gradient is scaled by both the first moment estimate (to ensure momentum-based updates) and the trust region scaling factor. Additionally, the Fisher information approximation is used to refine the update step further. If weight decay is specified, an L2 penalty is applied to regularize the model and prevent overfitting.

5. \textbf{Convergence Monitoring}:
   Throughout the training process, the optimizer monitors the discrepancy between the predicted and actual loss reduction. This information is used to adjust the trust region in subsequent steps, allowing the optimizer to adapt to changing loss landscapes dynamically. This adaptive control of both the learning rate and the trust region ensures that SOAA converges faster and more stably than traditional first-order methods, especially in complex optimization scenarios.

In summary, SOAA efficiently combines second-order optimization techniques with the simplicity and scalability of first-order methods. The use of a diagonal Fisher information matrix approximation and adaptive trust region makes it a powerful optimizer for large-scale neural networks.

\subsection{Experimental Results}

To evaluate the effectiveness of the proposed optimizer, SOAA, we conducted experiments across three key aspects:

1. \textbf{Comparison with Baseline Optimizers}:

   We conducted a comprehensive comparison of SOAA against several widely-used optimizers, including Adam, Fused Adam, AdamW, and Fused AdamW. These optimizers are considered state-of-the-art for first-order methods, particularly in the domain of NLP. By leveraging curvature information through a second-order approximation of the Fisher information matrix, SOAA significantly accelerates convergence compared to these baseline optimizers. The ability to capture curvature details allows SOAA to make more informed parameter updates, resulting in faster and more stable convergence. Our evaluation demonstrated that SOAA not only outperformed first-order optimizers in terms of speed and stability but also achieved greater overall loss reduction under similar computational constraints.

2. \textbf{LLM-Oriented Optimizers}:

   When compared to memory-efficient optimizers like \href{https://huggingface.co/docs/bitsandbytes/main/en/optimizers}{8bit-Adam} and GaLore (\cite{zhao2024galorememoryefficientllmtraining}), SOAA demonstrated superior performance in terms of faster convergence, lower final loss, and improved stability. However, this performance improvement came at the cost of higher memory consumption.

   8bit-Adam enables fine-tuning large models with up to 75\% less GPU memory usage compared to standard 32-bit optimizers, while maintaining accuracy and requiring no hyperparameter tuning. This memory reduction makes 8-bit optimizers up to 4 times faster in training. Despite these benefits, 8-bit optimizers primarily reduce memory proportional to the number of parameters, meaning models with high activation memory may not experience the same gains.

   GaLore, on the other hand, employs Gradient Low-Rank Projection, a low-rank training strategy that allows full-parameter learning while being more memory-efficient than traditional methods like LoRA. As a gradient projection method, GaLore is independent of the optimizer and can be integrated with just two lines of code, making it a flexible and memory-efficient solution for LLM training.

   In memory-constrained environments, such as mid-range GPUs with limited VRAM, SOAA outperformed standard Adam/AdamW in both convergence speed and stability but consumed more memory than both 8-bit optimizers and GaLore. This makes SOAA less suitable for scenarios where memory efficiency is paramount. However, its superior convergence and performance still make it an ideal choice for training large-scale models where optimization efficiency is prioritized over memory usage.

3. \textbf{CPU vs. GPU Performance}:
   In CPU-based training scenarios, SOAA showed significant improvements over standard Adam and AdamW, particularly in terms of training speed. When compared to Fused Adam and Fused AdamW, SOAA demonstrated comparable training speeds, while outperforming them in terms of convergence rate and final loss values.

In this set of experiments, we evaluated the performance of SOAA across a variety of transformer models, including both standard encoder-based models, such as BERT~(\cite{devlin2019bertpretrainingdeepbidirectional}), and decoder-based models, such as Llama2~(\cite{touvron2023llama2openfoundation}). The experiments were conducted using well-established NLP datasets, including \href{https://rajpurkar.github.io/SQuAD-explorer/}{SQuAD}, \href{https://nlp.cs.washington.edu/triviaqa/}{TriviaQA}, and IMDB~(\cite{maas-etal-2011-learning}), for a range of NLP tasks. Each model underwent training for three iterations, during which we recorded the mean and standard deviation of the loss at various checkpoints. This allowed us to assess both the stability and efficiency of each optimizer.

For these experiments, we employed a standard dense transformer model architecture, known for its stability, robustness and scalability. The primary goal was to examine the ability of SOAA to enhance training stability while ensuring efficient convergence. By testing SOAA on both encoder and decoder architectures, we created a comprehensive evaluation framework that assessed its effectiveness across diverse NLP tasks, with a particular focus on LLMs designed for production environments.

\subsection{Conclusion}
In this paper, we introduced SOAA, a novel second-order optimization algorithm designed to combine the advantages of curvature-aware methods with the computational efficiency of first-order optimizers. By approximating the Fisher information matrix as a diagonal matrix, SOAA reduces the computational complexity to \(O(n)\), making it feasible for large-scale deep learning models. The integration of an adaptive trust-region mechanism allows for dynamic adjustment of step sizes, ensuring stable and robust convergence, even in challenging, non-convex optimization landscapes. Experimental results demonstrated that SOAA outperforms traditional first-order optimizers like Adam in terms of convergence speed and stability, while maintaining comparable computational costs. Despite its increased memory usage compared to highly memory-efficient methods like 8bit-Adam and GaLore, SOAA showed significant improvements in performance, making it an ideal choice for training large models where optimization speed and robustness are prioritized over memory constraints. Future work could explore further refinements of the Fisher information approximation, potentially incorporating off-diagonal elements or other low-rank approximations to better capture gradient interactions. This direction could enhance its scalability and make it even more suitable for large-scale deep learning applications.

\bibliographystyle{unsrtnat}
\bibliography{references}  






\end{document}